\documentclass[a4paper]{article} 
\usepackage{spconf,amsmath,epsfig,graphics,url}
\usepackage[utf8]{inputenc}
\usepackage{amssymb}
\usepackage{algpseudocode}
\usepackage{algorithm}

\usepackage{fancyhdr}
\newcommand{\G}{\mathcal{G}} 
\newcommand{\E}{\mathcal{E}}
\newcommand{\V}{\mathcal{V}}
\newcommand{\W}{\mathbf{W}} 

\title{Graph-based denoising for time-varying point clouds}
\name{Yann~Schoenenberger, Johan~Paratte, Pierre~Vandergheynst}
\address{Signal Processing Laboratory (LTS2), Ecole Polytechnique Fédérale de Lausanne, 1015, Switzerland}

\begin{document}
\ninept
\urlstyle{same}
\maketitle
\begin{abstract}
Noisy 3D point clouds arise in many applications. They may be due to errors when constructing a 3D model from images or simply to imprecise depth sensors. Point clouds can be given geometrical structure using graphs created from the similarity information between points. This paper introduces a technique that uses this graph structure and convex optimization methods to denoise 3D point clouds. A short discussion presents how those methods naturally generalize to time-varying inputs such as 3D point cloud time series.
\end{abstract}


\section{Introduction}
\label{sec:intro}

Recovering 3D information from pictures or videos is a central topic in computer vision. Multi-view stereo techniques are the current state-of-the-art in this field. There are many algorithms that are able to create 3D models from images, see \cite{furukawa2010accurate}. The usual output of such methods are noisy 3D point clouds. We distinguish two types of noise. 

Firstly, a reconstructed point may be slightly off due to imprecise triangulation occurring in the 3D reconstruction algorithm. We refer to this as position noise. It is reasonable to assume that this noise is white Gaussian noise and thus averages to zero with high probability over a large number of spatial or temporal samples. 

Secondly, there may be points that are put at a completely wrong location depending on the details of the 3D reconstruction algorithms (e.g. a false epipolar match). They are nothing but very extreme cases of position noise. We propose to handle these outliers separately as they are not white Gaussian noise. Both position noise and outlier addition are presented in the context of 3D reconstruction, but they may be encountered in much more general settings like the output of depth cameras. 

The 3D point clouds we consider are by definition scattered sets of points. However, these points are not distributed randomly and their distribution follows an underlying structure as they describe 3D shapes in space. Moreover, these shapes usually possess a certain degree of smoothness or regularity. It is precisely this assumption of smoothness that will be used here for denoising. We assume that the point cloud is a sampled version of a set of smooth manifolds.

The underlying manifold can be approximated by creating a graph from the point cloud. Indeed, it has been shown that if a graph is constructed from points that are samples drawn from a manifold, the geometry of the graph is similar to the geometry of the manifold. In particular, the Laplacian of the graph converges to the Laplace-Beltrami operator of the manifold \cite{belkin2005towards}. 

Usually, point cloud denoising is done by estimating surface normals and averaging along the normal direction in small neighbourhoods of points, see e.g. \cite{mitra2004estimating} \cite{yagou2002mesh}, or using simple statistical methods \cite{rusu2008towards}. In this paper, we propose to use the graph created from the point cloud to tackle the problem of denoising using signal processing on graphs and convex optimization in particular. We first explain how to construct a graph from the points. Then we show how the positions can be interpreted as a graph signal which can be filtered and denoised using modern convex optimization methods. We also show that our method is general and extends naturally to point cloud time series. Finally, we show the effectiveness of those methods on real-world data sets and quantitatively assess their performance on synthetic point clouds. To the best of our knowledge, the approach presented here is the first to use signal processing and convex optimization on graphs for point cloud denoising.


\begin{figure}[t]
\includegraphics[width=\linewidth]{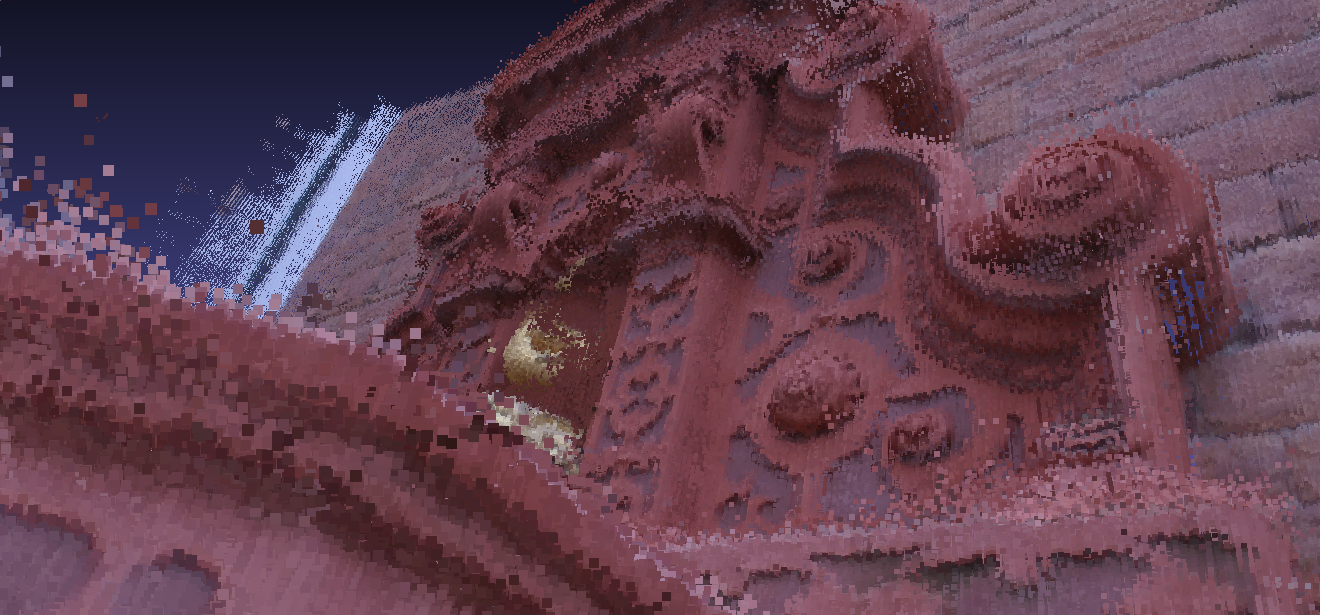}

\vspace{0.1cm}

\includegraphics[width=\linewidth]{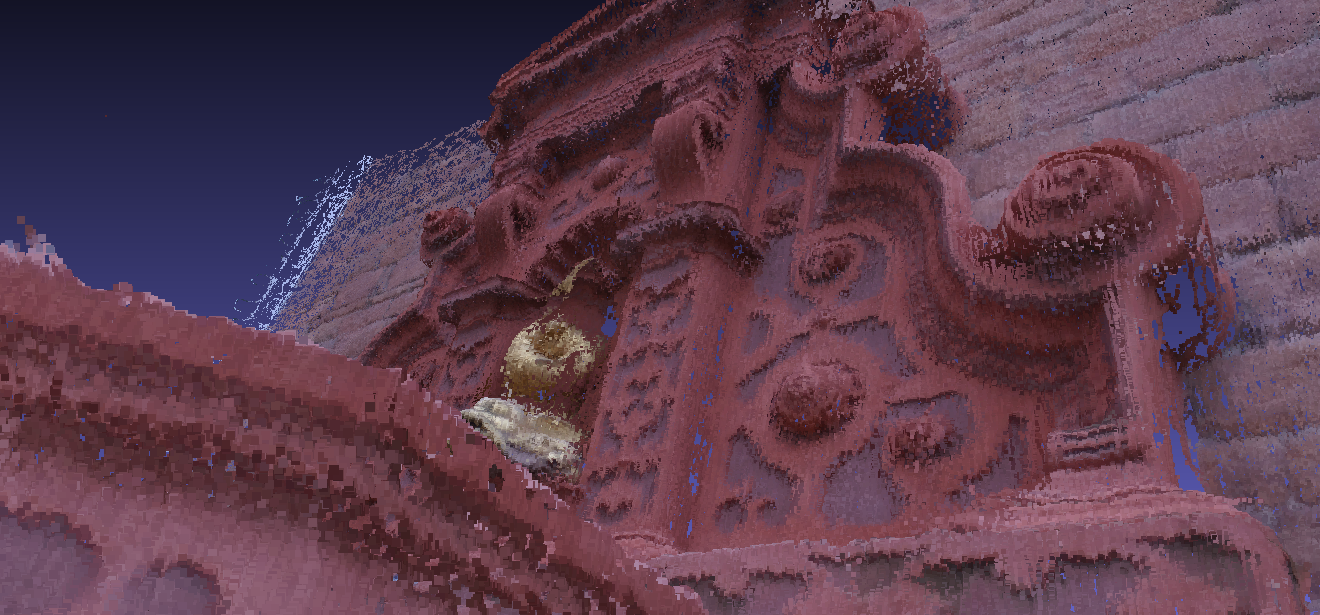}

\vspace{0.1cm}

\caption{Result of the proposed denoising process on a real-world example. \textbf{Top:} a noisy point cloud obtained using a state-of-the-art Multi-View Stereo reconstruction technique. \textbf{Bottom:} the point cloud obtained after processing with the proposed graph-based denoising methods. }
\label{fig:fountain}
\end{figure}

\section{Problem definition}
\label{sec:probdef}

\subsection{Graph nomenclature}
\label{ssec:grapn}
An weighted undirected graph $\G = (\V,\E, \W)$ is defined by a finite set of vertices $\V$ with $| \V | = N$, a set of edges $\E$ of the form $(v_i,v_j) \in \E$ with $v_i,v_j \in \V$ and a weighted adjacency matrix $\W$. The entry $W_{i,j} \in \mathbb{R}^{+}$ of $\W$ is the weight associated with the edge $(v_i,v_j)$ and $W_{i,j} = 0$ if and only if $(v_i,v_j) \notin \E$. Since only undirected graphs are considered, the matrix $\W$ is symmetric. 

To process data living on the graph, the notion of graph signals are needed. Given a graph $\G$, a graph signal (or function) is defined on the vertices of a graph $f : \V \rightarrow \mathbb{R}$. This is equivalent to a vector $\mathbf{f} \in \mathbb{R}^N$ where $\mathbf{f}=(f(v_1),\ldots,f(v_N))^T$. 

Processing such signals is done using graph signal processing techniques (see \cite{shuman2013emerging} for an introduction). One can apply many approaches which are graph-based equivalents of classical signal processing such as spectral analysis, filtering or convolution. In particular, methods to filter or denoise graph signals have been developed recently, see \cite{zhang2008graph}, \cite{smola2003kernels}.


\subsection{Graph construction from a point cloud}
\label{ssec:grapc}
In our case we are only given as input a point cloud denoted $P = \{p_1, p_2, \ldots, p_n\}$ with $\forall i:p_i \in \mathbb{R}^3$. We thus need a way to construct a graph given the point cloud. A standard way is to do a $k$-nearest neighbours ($k$-NN) construction as it makes geometric structure explicit, see \cite{wang2012geometric}. Every vertex is connected through an edge to its $k$ nearest neighbours with an associated weight which is computed given some metric. In this context we choose to use the Euclidean distance. A very standard choice for the weighting function is to use the thresholded Gaussian kernel 

\begin{eqnarray*}
\label{eq:weight}
 W_{i,j} = \left\{
  \begin{array}{l l}
    exp{\left(-\frac{||p_i-p_j||^2}{2\theta^2}\right)} & \quad \text{if $p_j \in C_{k}(i)$ or $p_i \in C_{k}(j)$ }\\
    0 & \quad \text{otherwise}
  \end{array} \right.
\end{eqnarray*}

In this equation, $\theta$ is a variance hyperparameter and $C_{k}(i)$ is the set containing the $k$ closest points to $p_i$. An alternative way to construct the graph is to connect each vertex to all its neighbours in a ball of radius $\epsilon$.

\subsection{Point cloud processing}
\label{ssec:pclproc}

Once a graph is constructed from a point cloud, we have a structure enforcing the geometrical shape defined by the set of points. Since we want to denoise the spatial coordinates of the points, the graph signal $f$ we consider is $3$-dimensional and defined by $f(v_i) = p_i$. Associating the 3D coordinates to each vertex allows us to measure the local smoothness of the point cloud using the smoothness of the graph. 

Note that the positions of the points are used both to construct the graph and as the signal to be processed. Thus, if the position of a point is modified, the structure of the graph needs to be updated. The $k$-NN graph corresponding to the processed point cloud will have different edges and edge weights than the $k$-NN graph of the original point cloud. The position denoising scheme presented in subsection~\ref{ssec:posden} can be made iterative by computing the $k$-NN graph of the output of the denoising procedure and running the denoising procedure again. However, as shown in subsection~\ref{ssec:perfev}, one iteration of the algorithm already yields very good results.

Many existing denoising methods such as \cite{yagou2002mesh}, \cite{vollmer1999improved} or \cite{wei2015bi} use meshes as input instead of point clouds. In those instances, the mesh can be seen as an approximation of an underlying manifold and denoising means smoothing this surface. Our method differs greatly in that it uses graph signal processing with only the point cloud as input. By directly working on the point cloud we avoid having to create a mesh which is a complex and error-prone process. 

\subsection{Position denoising}
\label{ssec:denoise}

Since the point cloud $P$ is noisy one can express each $p_i$ as the sum $p_i = x_i + n_i$ of the unknown true position $x_i$ and a noise term $n_i$ where $\forall i: p_i,x_i,n_i \in \mathbb{R}^3$. Ideally, one would like to recover $X = \{x_1, x_2, \ldots, x_n\}$ from $P$ perfectly, but this is not exactly what we aim for. Since, in our framework, a point cloud is a discrete sampling of a 2-dimensional manifold $M$ in 3-dimensional space, denoising means moving the points closer to (ideally on) $M$. Removing the noise from $p_i$ does not mean recovering $x_i$, but mapping it to a point on $M$ and the error is the shortest distance to a point on that manifold. 

Using the above definitions the graph $\G$, constructed from the point cloud $P$, can be seen as the discrete and noisy approximation of $M$. The smoothness of the coordinates signal $f$ on $\G$ is thus directly linked to the proximity of the points to the manifold $M$. The smoothness of $f$ on $\G$ can be measured using the graph gradient, see \cite{shuman2013emerging}. In section~\ref{sec:method} we propose convex optimization methods to enforce the smoothness of $f$ on $\G$ while keeping the points close to their original location. 

In practice, outliers (i.e. points that are very far away from their true position on $M$) should be removed altogether so they do not skew the position denoising. An outlier, by definition has very few close neighbours in the $k$-NN graph. Our algorithm takes advantage of this to choose which points to remove.

\section{Proposed method}
\label{sec:method}

\subsection{Algorithm for outlier removal}
\label{ssec:outrem}

The first step is to remove the outliers so they do not skew the position denoising algorithm. For this, we construct an $\epsilon$-NN graph from the point cloud. Because outliers are, by definition, very distant from inlier points, their degree, defined as the sum of the weights on all adjacent edges, will in average be significantly lower than that of inlier points. Thus, erroneous points can be eliminated by removing all vertices (and corresponding points) having a degree below some threshold $\tau$. This threshold is a parameter that can be set after the $\epsilon$-NN graph is computed such that a given percentile of outliers is removed. 


\subsection{Algorithm for position denoising}
\label{ssec:posden}

The second step is to correct the position of the remaining vertices. At this stage, no vertex is removed, but the locations are corrected. To do so, we consider a $k$-NN graph $\G$ constructed from the points remaining after outlier removal. We consider a graph signal $f$ defined as the spatial coordinates of each point as defined above. 

As already introduced, the problem of denoising a signal on the graph can be written as a convex minimization problem with the constraint that the denoised signal must be smooth on the graph. We write the optimization problem as

\begin{equation}
\label{eq:tik}
\dot{\mathbf{x}} = \operatorname*{arg\,min}_\mathbf{x} ||\mathbf{x}-\mathbf{f}||_2^2+\gamma||\nabla_G\mathbf{x}||_2^2 .
\end{equation}

In equation~\ref{eq:tik}, $\dot{\mathbf{x}}$ is the estimated denoised signal, $\mathbf{f}$ the noisy signal, $\gamma$ a regularization parameter and $\nabla_G\mathbf{x}$ the gradient of the signal $\mathbf{x}$ on the graph $\G$ as defined in \cite{shuman2013emerging}. The first term of the optimization is an energy term which constrains the denoised points to be close to their original positions. The second term is the smoothness constraint. The solution of this problem is shown to be a filtering on graph with filter $g(\lambda_\ell) = \frac{1}{1+2\gamma \lambda_\ell} $, see \cite{shuman2013emerging}.


The Tikhonov regularization presented in equation~\ref{eq:tik} can be replaced by a Total Variation ($TV$) regularization, if we assume the manifold underlying the point cloud to be piecewise smooth instead of smooth. With this new constraint, the convex optimization problem becomes

\begin{equation}
\label{eq:tv}
\dot{\mathbf{x}} = \operatorname*{arg\,min}_\mathbf{x} ||\mathbf{x}-\mathbf{f}||_2^2+\gamma||\nabla_G\mathbf{x}||_1.
\end{equation}

In equation~\ref{eq:tv}, the variables are the same as in equation~\ref{eq:tik}. This problem can be efficiently solved by the alternating direction method of multipliers (ADMM), see \cite{boyd2011distributed}.

\subsection{Extension to time-varying point clouds}
\label{ssec:highdim}

It is worth emphasizing that the presented algorithms can be applied to any data set with a meaningful distance function enabling the creation of $\epsilon$-NN and $k$-NN graphs. For example, in the case of a point cloud time series (e.g. created from a set of videos) rather than a static one (e.g. created from a set of pictures), it is possible to exploit temporal distance in addition to spatial distance in order to also enforce smoothness in time.

A scheme that we put in practice and that works well is for a given point $p$ at time $t$, $p$ is connected to its $k_1$ nearest neighbours at time $t$ as well as its $k_2$ nearest neighbours at time $t-1$ and its $k_2$ nearest neighbours at time $t+1$. Picking $k_1 > 2k_2$ seems to give good results. Of course, it is assumed that the coordinates system in use allows to meaningfully compute the distance between points from one time-step to the next. In practice, this is a very reasonable assumption.

\section{Experimental results}
\label{sec:expres}

Although the methods perform well on point cloud time-series, the focus of this section is the analysis of the experimental results in the case of static 3D point clouds which are easier to visualize. The algorithms have been implemented using the GSPBox \cite{perraudin2014gspbox} for the graph signal processing aspects and the UNLocBoX \cite{perraudin2014unlocbox} for convex optimization. All results can be reproduced using free software and data available online~\footnote{\url{https://lts2.epfl.ch/research/reproducible-research/graph-based-point-cloud-denoising/}}.  

\subsection{Application to real data}
\label{ssec:app}

\begin{figure}[t]
\includegraphics[width=\linewidth]{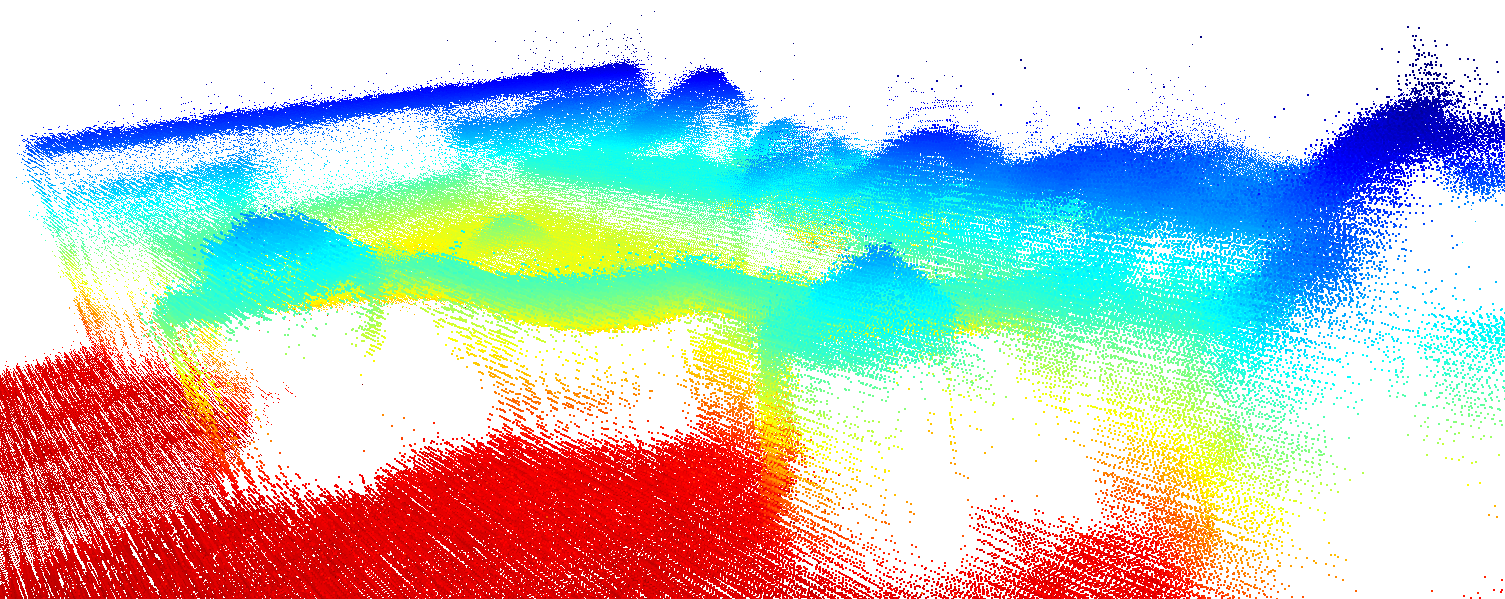}
\includegraphics[width=\linewidth]{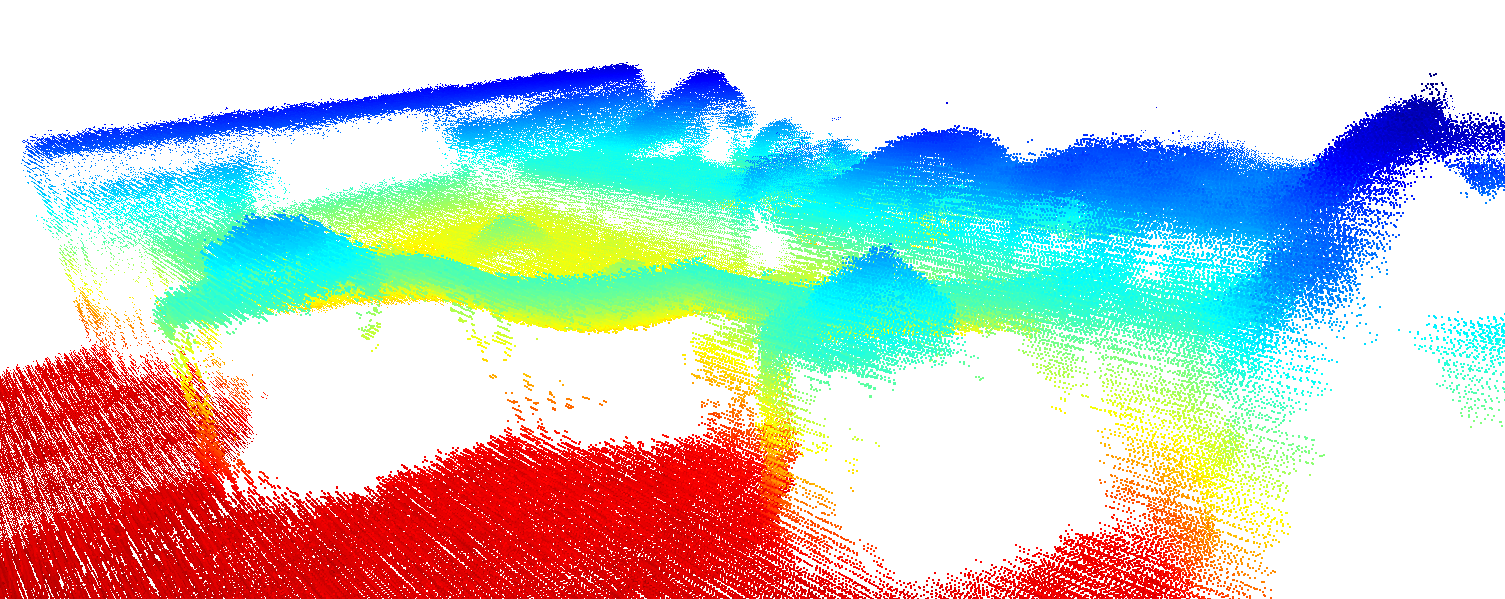}
\includegraphics[width=\linewidth]{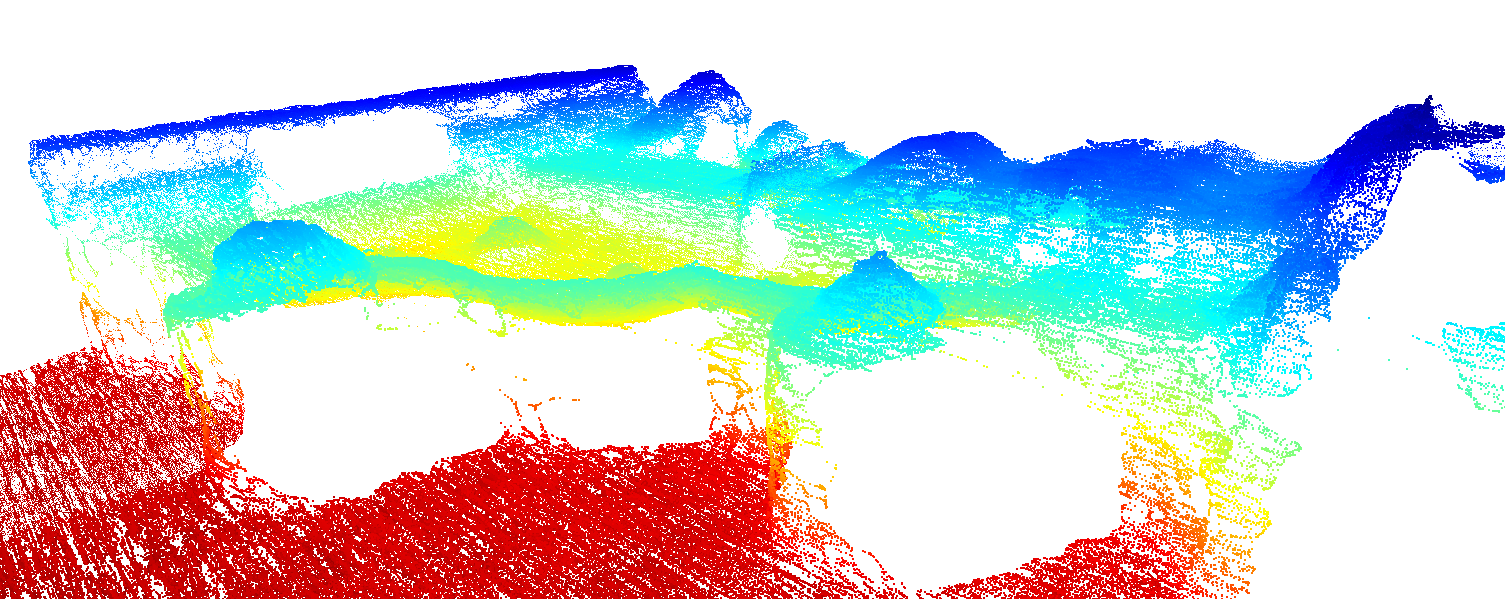}
\vspace{0.5em}
\caption{Denoising process of a point cloud (from real-world data) containing $1000662$ points. \textbf{Top:} original noisy point cloud. \textbf{Middle:} point cloud after degree filtering, with $941316$ points remaining. \textbf{Bottom:} resulting denoised point cloud after degree filtering and position denoising with the $TV$ constraint.}
\label{fig:basepc}
\end{figure}

We used a multi-view stereo algorithm to construct a point cloud from the fountain data set \cite{strecha2008benchmarking}. This point cloud was then denoised using the methods presented in this paper. We have found that $\epsilon = 0.01$ and $k=10$ for the graph construction and $\tau=3$ for outlier removal are good parameters for all of our experiments. Figure~\ref{fig:fountain} shows the noisy point cloud (top) and the denoised version (bottom). Figure~\ref{fig:basepc} (top) depicts the raw reconstructed point cloud where we can clearly observe noise and outliers. Figure~\ref{fig:basepc} (middle) shows the point cloud after degree filtering and we can observe that outliers have been removed. Finally, Figure~\ref{fig:basepc} (bottom) shows the point cloud after denoising using the $TV$ regularization constraint which is better suited for real-world data since it promotes piecewise smoothness rather than overall smoothness.  We can observe that the final point cloud is sharper. In addition the sampled 3D shapes are piecewise smoother: the edges have been preserved while the fluctuations in the positions of the points (due to noise) have been reduced. The color is based on the depth, which allows a better visual inspection than the true colors. 


\subsection{Performance evaluation}
\label{ssec:perfev}

\begin{figure*}[ht]
\includegraphics[width=\linewidth]{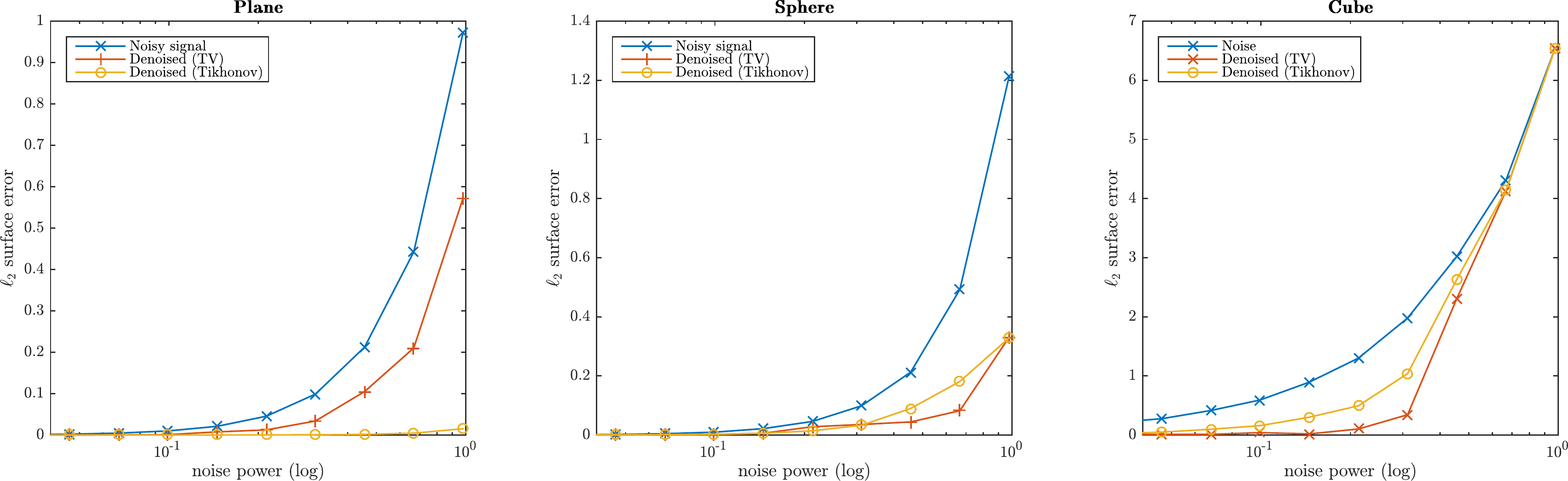}
\caption{Average error before and after position denoising of three synthetic point clouds: a square plane (\textbf{left}), a sphere (\textbf{center}) and a cube (\textbf{right}). All underlying manifolds are known and were sampled uniformly at random $10000$ times each.}
\label{fig:errorssigma}
\end{figure*}

Figure~\ref{fig:fountain} and \ref{fig:basepc} as well as subsection~\ref{ssec:app} show that on real data, the denoising is of very good quality. However, we also would like to have a quantitative assessment of the denoising. 
To be able to measure the noise as defined in subsection~\ref{ssec:denoise}, we present an evaluation done on synthetic data sets so that the analytic form of the sampled manifold is known. The difficulty of doing it on real-world data lies in the fact that measuring the error requires a ground-truth manifold which is a continuous object very difficult to capture in practice. 

The chosen shapes are a sampled sphere (smooth), a sampled cube (piecewise smooth) and a sampled plane. 
The results of the experiments can be found on figure~\ref{fig:errorssigma}. 

On each one of the graphs presented in figure~\ref{fig:errorssigma}, the average distance of the points in the output point cloud from the ground-truth manifold is shown for nine different input noise levels. Note the logarithmic scale for the input noise power. There are three data series shown on each graph. The blue (crosses) points correspond to the noisy point cloud before any processing. The red (vertical crosses) points correspond to the output point cloud after position denoising with $TV$ regularization (shown in equation~\ref{eq:tv}). The yellow (circles) points correspond to the output point cloud after position denoising with Tikhonov regularization (shown in equation~\ref{eq:tik}). Note that using outlier removal before position denoising can only improve those results, but this is not shown since the focus is on the contribution of the graph-based position denoising method.

In the case of the square plane, a very simple and smooth surface, we see that Tikhonov regularization gives excellent results even with a lot of noise in the input. Using $TV$ regularization also allows for some denoising but less so. The fact that Tikhonov regularization outperforms $TV$ regularization on the plane is easily explained since the $TV$ prior will tend to leave a few discontinuities in the signal. For more complex shapes (the sphere and the cube), both $TV$ and Tikhonov regularizations yield good results with $TV$ slightly outperforming Tikhonov in most cases. With very high noise levels on shapes such as the cube which has sharp edges, the proposed denoising method breaks down and does not remove much noise. This intuitively makes sense, because as noise increases, sharp edges and other such features get blurred to a point where they disappear completely. Those cases are useful to asses that algorithms are well-behaved even under extreme circumstances, but they do not arise often in practice. 


\section{Conclusion and Future Work}
\label{sec:conc}

In this paper, we proposed point cloud denoising methods based on graph signal processing. We then showed how the algorithms perform on a real-world example and quantitatively assessed the performance of the methods using synthetic point clouds.

Our method is very general and can be extended to higher-dimensional spaces, including but not limited to time-varying point clouds (e.g. point clouds reconstructed from a set of videos rather than static images). We have argued how our proposed methods can be applied to those cases. It would be a good extension of the work presented here to do an in-depth analysis of the very promising results we obtained on various higher-dimensional extensions including various methods of creating graphs. 

\section{Acknowledgement}
\label{sec:ack}

The work presented in this paper has partially been done in the context of the SceneNet project. This project is funded by the European Union under the 7th Research Framework, programme FET-Open SME, Grant agreement no. 309169.


\bibliographystyle{IEEEbib}
\bibliography{refs}

\end{document}